%% file: main.tex
\pdfoutput=1

\documentclass[11pt]{article}

\usepackage{acl}

\usepackage{times}
\usepackage{latexsym}
\usepackage{multirow}
\usepackage{booktabs}
\usepackage{natbib}
\usepackage{titlesec}
\usepackage[ruled,vlined]{algorithm2e}
\usepackage[utf8]{inputenc}
\usepackage{amssymb,amsmath,caption}
\usepackage[hang,flushmargin]{footmisc}
\usepackage{lipsum}
\usepackage{listings}
\usepackage{extarrows}
\usepackage{subfigure}
\usepackage{xspace}
\usepackage{graphicx}
\usepackage{makecell}
\usepackage{ifthen}
\usepackage{xifthen}
\usepackage{wrapfig}
\usepackage{amsthm}
\usepackage{threeparttable}
\usepackage{microtype}
\usepackage{colortbl}
\usepackage{makecell}
\usepackage{xspace}

\usepackage[T1]{fontenc}

\usepackage[utf8]{inputenc}

\usepackage{microtype}
\usepackage{colortbl}
\usepackage{makecell}
\usepackage{threeparttable}
\usepackage{amssymb}
\usepackage{pifont}

\newcommand{\vpara}[1]{\vspace{0.07in}\noindent\textbf{#1}\xspace}

\newcommand{\hide}[1]{} 
\newcommand{\tabincell}[2]{\begin{tabular}{@{}#1@{}}#2\end{tabular}}

\newcommand\todo[1]{\textbf{\textcolor{orange}{[TODO]}}}
\newcommand\ours{P-Tuning\xspace}

\newcommand{\cmark}{\ding{51}}%
\newcommand{\xmark}{\ding{55}}%

\newcommand\std{\scriptsize$\pm$}
\usepackage{enumitem}
\setenumerate[1]{itemsep=0pt,partopsep=0pt,parsep=\parskip,topsep=2pt,leftmargin=0.4cm}
\setitemize[1]{itemsep=0pt,partopsep=0pt,parsep=\parskip,topsep=2pt,leftmargin=0.4cm}
\setdescription{itemsep=0pt,partopsep=0pt,parsep=\parskip,topsep=2pt,leftmargin=0.4cm}

\newcommand{\solution}[0]{P-tuning\xspace}

\title{GPT Understands, Too}

\author{
  Xiao Liu$^{1*}$, 
  Yanan Zheng$^{1*}$, 
  Zhengxiao Du$^{1}$, 
  Ming Ding$^{1}$, 
  Yujie Qian$^{2}$,\\
  \textbf{Zhilin Yang$^{1\dagger}$, 
  Jie Tang$^{1\dagger}$} \\
  $^1$Tsinghua University \qquad
  $^2$Massachusetts Institute of Technology}

\begin{document}
\maketitle

\let\thefootnote\relax\footnotetext{
$^\dagger$ corresponding to: Zhilin Yang (zhiliny@tsinghua.edu.cn) and Jie Tang (jietang@tsinghua.edu.cn) }{
\let\thefootnote\relax\footnotetext{
$^*$ indicates equal contribution.}
}

\begin{abstract}
    \input{0_abstract}
\end{abstract}

\input{1_intro}
\input{4_method}

\input{5_exp}
\input{2_related}
\input{6_conclusion}

\bibliography{anthology,custom}
\bibliographystyle{acl_natbib}


\end{document}

%% file: 0_abstract.tex

Prompting a pretrained language model with natural language patterns has been proved effective for natural language understanding (NLU). However, 
our preliminary study reveals that manual discrete prompts often lead to unstable performance---e.g., changing a single word in the prompt might result in substantial performance drop. We propose a novel method \ours that employs trainable continuous prompt embeddings in concatenation with discrete prompts. Empirically, P-Tuning not only stabilizes training by minimizing the gap between various discrete prompts, but also improves performance by a sizeable margin on a wide range of NLU tasks including LAMA and SuperGLUE. \ours is generally effective for both frozen and tuned language models, under both the fully-supervised and few-shot settings.

%% file: 1_intro.tex
\section{Introduction}

Pretrained language models~\cite[PLMs;][]{brown2020language} have significantly advanced the performance of natural language understanding (NLU).
PLMs are trained with different pretraining objectives, such as masked language modeling \cite{devlin2018bert}, autoregressive language modeling \cite{radford2019language}, seq2seq \cite{raffel2019exploring}, and permutation language modeling \cite{yang2019xlnet}.
PLMs can be further enhanced with prompting \cite{brown2020language,schick2020small}, which employs manually written prompt patterns as additional input to a language model.
With prompting while PLMs are either finetuned on a small labeled dataset or frozen for direct inference on downstream tasks. Prompting has significantly improved the performance of many NLU tasks \cite{brown2020language,schick2020small}.

However, we observe that manual discrete prompts suffer from a large degree of instability. As shown in Table \ref{tab:motivation_lama}, with a frozen language model, changing a single word in the prompt might result in substantial performance drop. As we will show in Section \ref{sec:exp}, when the language model is tuned, the instability problem is alleviated but the performance difference between different prompts is still sizeable, especially in the few-shot setting. Such an instability issue of discrete prompts poses a critical challenge in practice. Recent approaches of automatic prompting have attempted to search for a better-performing prompt given a task \cite{shin2020autoprompt,gao2020making,jiang2020can}, but these methods do not change the unstable nature of discrete prompts.

\begin{figure}[t]
    \centering
    \includegraphics[width=\linewidth]{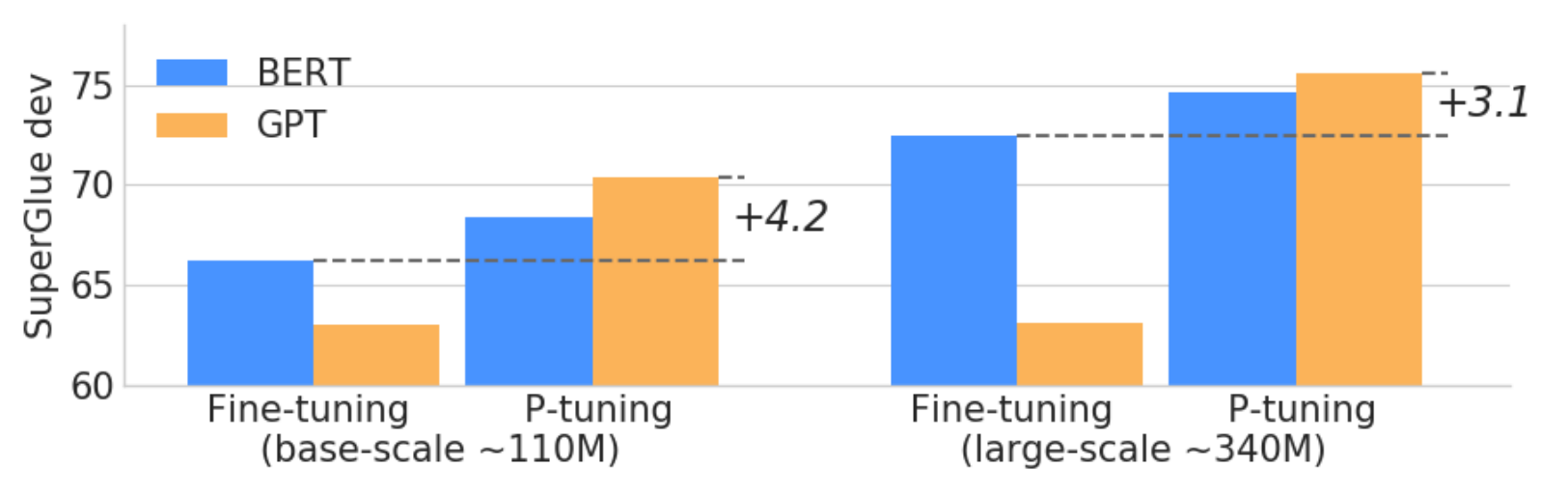}
    \caption{Average scores on 7 dev datasets of SuperGLUE using P-Tuning.}
    \label{fig:example}
\end{figure}



\begin{table}[t]
    \small
    \centering
    \renewcommand\tabcolsep{1.0pt}
    \begin{tabular}{lp{1.0cm}p{1.0cm}<{\centering}}
       \toprule[1pt]
       Prompt & P@1 w/o PT & P@1 ~w/ PT\\ \midrule[1pt]
       [X] is located in [Y]. \textit{(original)} & 31.3 & 57.8 \\ \hline
       [X] is located in which country or state? [Y]. & 19.8 & 57.8\\ \hline
       [X] is located in which country? [Y]. & 31.4 & 58.1 \\ \hline
       [X] is located in which country? In [Y]. & 51.1 & 58.1 \\ 
       \toprule[1pt]
    \end{tabular}
    \caption{Discrete prompts suffer from instability (high variance), while P-Tuning stabilizes and improves performance. Results are precision@1 on LAMA-TREx P17 with BERT-base-cased. ``PT'' refers to P-Tuning, which trains additional continuous prompts in concatenation with discrete prompts.} 
    \label{tab:motivation_lama}
    \vspace{-5mm}
\end{table}

To reduce the instability of discrete prompts, we propose a novel method \ours that employs trainable continuous prompt embeddings in concatenation with discrete prompts. Specifically, given a discrete prompt as the input, \ours concatenates continuous prompt embeddings with the discrete prompt tokens and feeds them as the input to the language model. The continuous prompts are updated by backpropagation to optimize the task objective. The intuition is that continuous prompts incorporate a certain degree of learnability into the input, which may learn to offset the effects of minor changes in discrete prompts to improve training stability. To further improve performance, we employ a prompt encoder using LSTMs or MLPs to model the dependency between continuous prompt embeddings.


We experiment with two NLU benchmarks: the LAMA~\cite{petroni2019language} knowledge probing and SuperGLUE~\cite{wang2019superglue}. 
On LAMA, with the language model frozen, \ours outperforms manual discrete prompts and searched prompts by 20+ points and 9 points respectively with the same pretrained models. On SuperGLUE, with the language model finetuned, \ours outperforms PET \cite{schick2020small} with the best discrete prompts under both the fully-supervised and few-shot settings. In addition to improving performance, our results show that across a wide range of tasks and settings, \ours substantially reduces the performance gap between different discrete prompts, which results in improved stability for language model adaptation.

%% file: 4_method.tex
\section{Method} \label{sec:method}

\input{3_motivation}

\begin{figure*}
    \centering
    \includegraphics[width=\linewidth]{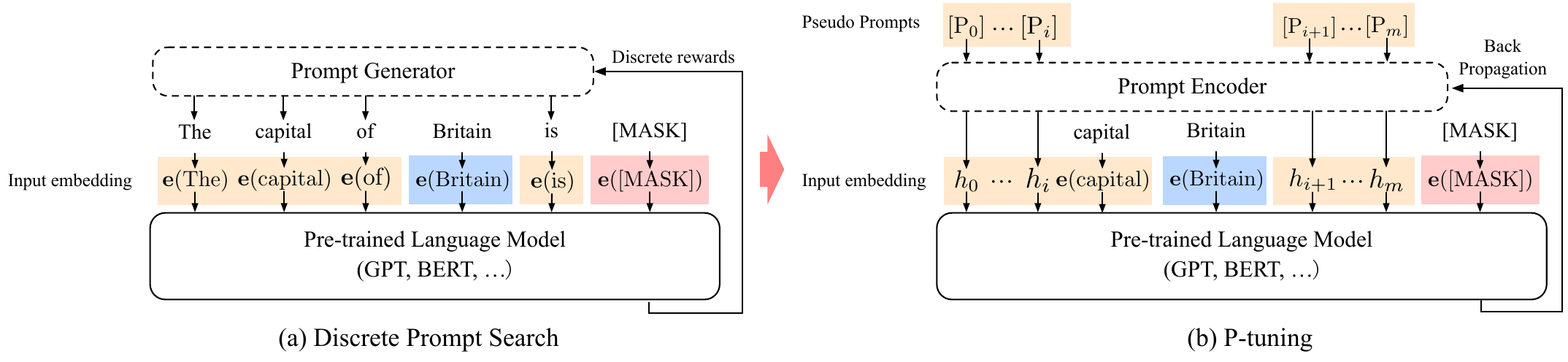}
    \caption{An example of prompt search for ``The capital of Britain is [MASK]''. Given the context (blue zone, ``Britain'') and target (red zone, ``[MASK]''), the orange zone refer to the prompt. In (a), the prompt generator only receives discrete rewards; on the contrary, in (b) the continuous prompt embeddings and prompt encoder can be optimized in a differentiable way.}
    \label{fig:framework}
\end{figure*}

\subsection{\ours}


Formally, let $\mathcal{M}$ be a pretrained language model with a hidden size of $h$ and a vocabulary size of $|\mathcal{V}|$.
Let $\{(\mathbf{x}_i, \mathbf{y}_i))\}_i$ be a labeled dataset for an NLU task, where $\mathbf{x}_{0:n}=\{x_0, x_1, ..., x_n\}$ is an input consisting of a sequence of discrete tokens, and $\mathbf{y} \in \mathcal{Y}$ is a label.
Our goal is to estimate the conditional probability for classification $f_{\mathcal{M}}(x) = \hat{p}(y | x)$ with parameters of $\mathcal{M}$ either finetuned or frozen.

Prompting was proposed in the format of discrete tokens~\cite{schick2020small}.
Let $[\mathrm{D}_i]$ be a discrete prompt token.
Each prompt can be described as a template $T=\{[\mathrm{D}_{0:i}], \mathbf{x} , [\mathrm{D}_{(i+1):j}] , \mathbf{y}, [\mathrm{D}_{(j+1):k}]\}$, which could organize the labeled data (including the inputs $\mathbf{x}$ and the label $\mathbf{y}$) into a sequence of text tokens, such that the task could be reformulated as filling in the blanks of the input text.
For example, for the task of predicting a country's capital (LAMA-TREx P36), a prompt could be ``The capital of [INPUT] is [LABEL].'' 
With a piece of labeled data ``(Britain, London)'', the reformulated text would be ``The capital of Britain is [MASK].'', where ``[MASK]" should predict the given label ``London''.
Both discrete prompts and discrete data are together mapped into input embeddings:
\[
\{\mathbf{e}(D_0)...\mathbf{e}(D_i), \mathbf{e}(x_0), ...,\mathbf{e}(x_n),...,\mathbf{e}(D_k)\}
\]
through the pretrained embedding layer, where $\mathbf{e}\in\mathbb{R}^{|\mathcal{V}|\times d}$.

However, as is discussed in Section~\ref{sec:issueswithdiscrete}, such discrete prompts tend to be extremely unstable and might not be optimal with back-propagation.
Therefore, we propose \ours that uses continuous prompt embeddings to improve and stabilize prompting.
Let [$\mathrm{P}_i$] be the $i^{\mathrm{th}}$ continuous prompt embedding.
The prompt template for \ours is as follows:
\[
T=\{[\mathrm{P}_{0:i}],\mathbf{x},[\mathrm{P}_{(i+1):j}],\mathbf{y},[\mathrm{P}_{(j+1):k}]\}
\]
\ours leverages an extra embedding function $f: [\mathrm{P}_i]\to h_i$ to map the template to
\[
    \{h_0,...,h_i, \mathbf{e}(\mathbf{x}),h_{i+1},...,h_j, \mathbf{e}(\mathbf{y}), h_{j+1},...,h_k\}
\]
Finally, we update the embeddings $\{P_i\}_{i=1}^k$ to optimize a task loss function.

It is noteworthy that we can also concatenate discrete prompts with continuous prompts, which performs better and is adopted throughout our experiments.
\ours is applicable to both frozen and finetuned language models.

\subsection{Prompt Encoder}

In the aforementioned framework, we employ a mapping function $f$ to map trainable embeddings $\{P_i\}$ to model inputs $\{h_i\}$. The intuition is that by using a mapping function, it is more convenient to model the dependency between different prompt embeddings, compared to using independent learnable embeddings. In our implementation, we use a lightweight neural network to formulate the function $f$. Specifically, we experiment with using long short-term memory (LSTM) networks, multi-layer perceptrons (MLPs), and the identity mapping function in Section \ref{sec:exp}.

%% file: 3_motivation.tex
\subsection{Issues with Discrete Prompts}\label{sec:issueswithdiscrete}

Prompting employs natural language patterns as additional inputs to pretrained language models for adaptation to downstream tasks \cite{brown2020language,schick2020small}.
Prior work~\cite{zheng2021fewnlu} has pointed out that prompting has achieved consistent and substantial improvements on a number of NLP tasks.
However, it still remains a challenging problem of how to write high-performing discrete prompts.

We performed preliminary experiments using different manual prompts on the LAMA knowledge probing task \cite{petroni2019language}, which aims to extract triplet knowledge from a language model by predicting the tail entities.
Results in Table~\ref{tab:motivation_lama} show that
manual discrete prompts lead to unstable performance. For example, if we compare the last two prompts in the table, changing a single word in prompt causes a drastic decrease of 20 points in performance.

In light of the challenge, recent works propose to automate the search procedure of discrete prompts by mining the training corpus~\cite{jiang2020can}, gradient-based searching~\cite{shin2020autoprompt}, and using pretrained generative models~\cite{gao2020making}. However, these works aim at searching for better-performing prompts but do not change the nature of instability for discrete prompts.
In addition to the instability issue, searching in the discrete space might not be able to fully leverage the gradients from backpropagation, which will potentially result in suboptimal solutions.
To this end, we explore the possibility of training continuous prompts to stabilize and improve the performance of language model adaptation.





%% file: 5_exp.tex
\section{Experiments} \label{sec:exp}


We include two NLU benchmarks: LAMA~\cite{petroni2019language} for knowledge probing (\textsection~\ref{sec:knoweledge}) and SuperGLUE~\cite{wang2019superglue} for general natural language understanding.
On SuperGLUE, we consider both the fully-supervised learning (\textsection~\ref{sec:fullysupervised}) and few-shot learning (\textsection~\ref{sec:few-shot}) settings.

On LAMA, following~\citet{shin2020autoprompt,jiang2020can}, language models are frozen and only the discrete or continious prompts are tuned.
For SuperGLUE, following \citet{schick2020small,zheng2021fewnlu}, language models are tuned. In our setting, we jointly optimize the language model parameters and the continuous prompts.
This setup not only follows the common, standard settings in prior work, but also allows evaluating P-Tuning with both tuned and frozen language models.

The overall task setup and a summary of results are shown in Table \ref{tab:summary}.

\begin{table}[]
\centering
\resizebox{0.76\linewidth}{!}{
\begin{tabular}{@{}lccc@{}}
\toprule
           & LAMA   & Full SG & Few SG \\ \midrule
 & frozen & tuned       & tuned        \\ \midrule
Improved   & \cmark & \cmark       & \cmark        \\
Stabilized & \cmark & \xmark       & \cmark        \\ \bottomrule
\end{tabular}}
\caption{Task settings and summary of results in our experiments. \solution shows improvement over baselines on all task settings, and can stabilize performance on LAMA and Few SG. For Full SG, the gap between discrete prompts is not large and training is stable even without \ours. (Full SG: fully-supervised learning on SuperGLUE; Few SG: few-shot SuperGLUE; Improved: overall performance improved; Stabilized: training stabilized by minimizing difference between discrete prompts).}
\label{tab:summary}
\end{table}

\input{5_1_LAMA}

\input{5_2_FullySup}
\input{5_3_FewShot}

%% file: 5_1_LAMA.tex
\input{tables/lama}

\subsection{Knowledge Probing}\label{sec:knoweledge}
\subsubsection{Setup}
Knowledge probing, or referred to as fact retrieval, evaluates how much real-world knowledge has language models gained from pre-training. The LAMA~\cite{petroni2019language} dataset evaluates it with cloze tests created from triples selected in the knowledge bases. 

\vpara{Datasets and vocabulary.} 
LAMA enforces all answers in single-token format. We first adopt the original LAMA-TREx dataset, consisting of 41 Wikidata relations and altogether 34,039 testing triples (namely LAMA-34k, which covers all BERT vocabularies). 
Since different pretrained models share distinct vocabularies, to allow direct comparison, we follow previous work \cite{shin2020autoprompt} to adopt a subset that covers the intersection of GPT's and BERT's vocabularies. This is caled LAMA-29k.
We again follow \citet{shin2020autoprompt} to construct the training, development, and test data to allow for fair comparison.


\vpara{Setup.}
LAMA has provided a handcraft prompt for each relation, as shown in Table \ref{tab:motivation_lama}, which are effective but likely sub-optimal. For bidirectional masked language models, we only need to replace ``[X]'' with the subject entity and ``[Y]'' with the [MASK] token; for unidirectional language models such as GPT, following LAMA's original setting on Transformer-XL~\cite{dai2019transformer}, we use the network output just before the target position. 

The number of prompt tokens and positions are selected based on the development sets, and for simplicity we choose the (3, sub, org\_prompt, 3, obj, 3) template for bidirectional models and (3, sub, org\_prompt, 3, obj) for unidirectional models as this configuration performs well for most relations (where the number indicates the number of continuous prompt tokens). Continuous prompts are concatenated with original discrete prompts.
During the prompt training, we set the learning rate to 1e-5 and use the Adam optimizer.

\subsubsection{Main results}
The results are presented in Table \ref{tab:lama_results}. \solution significantly improves the best results of knowledge probing from 43.3\% to 50.6\% on LAMA-34k and from 45.2\% to 64.2\% on LAMA-29k. Moreover, \solution outperforms previous discrete prompt searching approaches such as AutoPrompt~\cite{shin2020autoprompt} and LPAQA~\cite{jiang2020can} on the same-size models. This confirms our intuition in Section \ref{sec:method} that discrete prompts might not be optimal.

%% file: tables/lama.tex
\begin{table*}[tb]
\scriptsize
\begin{subtable}
    \centering
    \resizebox{0.47\linewidth}{!}{
    \begin{tabular}{c|l|c}
        \toprule[1pt]
        Prompt type                 & Model                                 & P@1  \\ 
        \midrule
        \multirow{3}*{\makecell[c]{Original\\(MP)}} & BERT-base             & 31.1 \\ 
        ~                           & BERT-large                            & 32.3 \\ 
        ~                           & E-BERT                                & 36.2 \\ 
        \midrule
        \multirow{3}*{Discrete}     & LPAQA \tiny{(BERT-base)}              & 34.1 \\
        ~                           & LPAQA \tiny{(BERT-large)}             & 39.4 \\
        ~                           & AutoPrompt \tiny{(BERT-base)}         & \underline{43.3} \\
        \midrule
        \multirow{2}*{\solution}    & BERT-base                             & 48.3 \\
        ~                           & BERT-large                            & \textbf{50.6} \\
        \bottomrule[1pt]
    \end{tabular}}
\end{subtable}%
\begin{subtable}
    \centering
    \resizebox{0.50\linewidth}{!}{
    \begin{threeparttable}
    \begin{tabular}{l|c|l}
        \toprule[1pt]
        Model               & MP                    & \solution \\ 
        \midrule 
        BERT-base (109M)    & 31.7                  & 52.3 \tiny{\color{red}{(+20.6)}}      \\ 
        \quad-AutoPrompt~\cite{shin2020autoprompt}  & -         & 45.2      \\ 
        BERT-large (335M)   & 33.5                  & 54.6 \tiny{\color{red}{(+21.1)}}     \\ 
        \midrule
        RoBERTa-base (125M) & 18.4                  & 49.3 \tiny{\color{red}{(+30.9)}}      \\ 
        \quad-AutoPrompt~\cite{shin2020autoprompt}  & -         & 40.0      \\ 
        RoBERTa-large (355M)& 22.1                  & 53.5 \tiny{\color{red}{(+31.4)}}     \\ 
        \midrule
        GPT2-medium (345M)  & 20.3                  & 46.5 \tiny{\color{red}{(+26.2)}}     \\ 
        GPT2-xl (1.5B)      & 22.8                  & 54.4 \tiny{\color{red}{(+31.6)}}     \\ 
        MegatronLM (11B)    & 23.1                  & \textbf{64.2} \tiny{\color{red}{(+41.1)}}     \\
        \bottomrule[1pt]
    \end{tabular}
    \end{threeparttable}}
    \end{subtable}
    
    \caption{Knowledge probing Precision@1 on LAMA-34k (left) and LAMA-29k (right). \solution outperforms all the discrete prompt searching baselines. (MP: Manual prompt; PT: \solution).}
\label{tab:lama_results}
\end{table*}

%% file: 5_2_FullySup.tex

\subsection{Fully-supervised Learning}\label{sec:fullysupervised}
\input{tables/full_superglue}
\subsubsection{Setup}

\vpara{Dataset.}
To evaluate \solution on fully-supervised learning tasks,
we adopt the SuperGLUE benchmark~\cite{SuperGLUE2019}, consisting of 8 challenging natural language understanding (NLU) tasks.
We focus on 7 of them since the ReCoRD~\cite{ReCoRD2018} task adopts no discrete prompts, thus \solution is not directly applicable.
The tasks include question answering (BoolQ~\cite{BoolQ2019} \& MultiRC~\cite{MultiRC2018}), textual entailment (CB~\cite{de2019commitmentbank} \& RTE~\cite{RTE2005}), co-reference resolution (WiC~\cite{wic-paper}), causal reasoning (COPA~\cite{COPA2011}), and word sense disambiguation (WSC~\cite{WSC2012}).

\vpara{Comparison methods.}
We experiment with \solution on both unidirectional and bidirectional pretrained models, i.e., GPT and BERT.
We include four variants BERT-Base, BERT-Large, GPT2-Base, and GPT-medium.
For each model, we compare standard classification finetuning, PET~\cite{schick2020small} (a typical finetuning method based on manual discrete prompts) and our \solution.

\vpara{Configuration.}
We use the same metrics as in~\cite{SuperGLUE2019}.
For fully-supervised learning, we use a large training set to finetune pretrained models and use a development set for hyper-parameter and model selection.
Specifically, 
the AdamW optimizer with a linearly decayed learning rate is used for training.
We use a learning rate of $\{1e-5, 2e-5, 3e-5\}$, a batch size of $\{16, 32\}$, and a warm-up ratio of $\{0.0, 0.05, 0.1\}$.
For small datasets (i.e., COPA, WSC, CB, RTE), we fine-tune pretrained models for 20 epochs.  
For larger datasets (i.e., WiC, BoolQ, MultiRC), we reduce the number of training epochs to be 10 as the model converges earlier.
Early stopping is used to avoid over-fitting the training data. 



\subsubsection{Main Results}
The main results of fully-supervised learning are shown in Table~\ref{tab:fully-supervised}.
We observe that \solution can improve fully-supervised learning performance on both BERTs and GPTs.
(1) Specifically, on the BERT-Base model, \solution achieves best performance on 5/7 tasks, while with BERT-Large, \solution outperforms other methods on 4/7 tasks.
The exceptions are WiC and MultiRC, both of which have relatively large training sets.
We find that \solution might not have large gains over CLS-FT on such high-resource tasks, while benefits more on low-resource tasks.
On average, \solution improves over the considered baselines.
(2) On GPT2-Base and GPT2-Medium models, \solution consistently achieves the best performance on all tasks.


%% file: tables/full_superglue.tex
\begin{table*}[tb]
\small
\centering
\subtable[Fully-supervised performance with base-scale models.]{
\resizebox{\linewidth}{!}{
    \begin{tabular}{l|c|cccccccccc}
    \toprule[1pt]
    & \multirow{2}{*}{\quad Method \quad} 
        & BoolQ 
        & \multicolumn{2}{c}{CB} 
        & WiC 
        & RTE 
        & \multicolumn{2}{c}{MultiRC}  
        & WSC 
        & COPA 
        & \multirow{2}{*}{Avg.} \\ 
    & & (Acc.) 
    & (Acc.) 
    & (F1) 
    & (Acc.) 
    & (Acc.) 
    & (EM) 
    & (F1a) 
    & (Acc.) 
    & (Acc.) \\ 
    \midrule
    \multirow{3}*{\shortstack{BERT-Base \\ (109M)}}
    & CLS-FT 
        & 72.9 & 85.1 & 73.9 & 71.1 & 68.4 & 16.2 & 66.3 & 63.5 & 67.0 & 66.2 \\	
    & PET-FT
        & 73.7 & 87.5 & 90.8 & 67.9 & 70.4 & 13.7 & 62.5 & 60.6 & 70.0 & 67.1 \\
    & \solution 
        & 73.9 & 89.2 & 92.1 & 68.8 & 71.1 & 14.8 & 63.3 & 63.5 & 72.0 & 68.4 \\
    \midrule[1pt]
    \multirow{3}*{\shortstack{GPT2-Base \\ (117M)}}
    & CLS-FT
        & 71.2 & 78.6 & 55.8 & 65.5 & 67.8 & 17.4 & 65.8 & 63.0 & 64.4 & 63.0\\
    & PET-FT 
        & 74.8 & 87.5 & 88.1 & 68.0 & 70.0 & 23.5 & 69.7 & 66.3 & 78.0 & 70.2 \\
    & \solution
        & 75.0 
        & 91.1 
        & 93.2 
        & 68.3 
        & 70.8 
        & 23.5 
        & 69.8 
        & 63.5 
        & 76.0 
        & 70.4 
        \\
    \bottomrule[1pt]
    \end{tabular}
}}
\subtable[Fully-supervised performance with large-scale models.]{
\begin{threeparttable}
    \resizebox{\linewidth}{!}{
    \begin{tabular}{l|c|cccccccccc}
    \toprule[1pt]
    & \multirow{2}{*}{\quad Method \quad} 
        & BoolQ & \multicolumn{2}{c}{CB} 
        & WiC 
        & RTE 
        & \multicolumn{2}{c}{MultiRC}  
        & WSC 
        & COPA 
        &\multirow{2}{*}{Avg.} \\ 
    & & (Acc.) 
    & (Acc.) 
    & (F1) 
    & (Acc.) 
    & (Acc.) 
    & (EM) 
    & (F1a) 
    & (Acc.) 
    & (Acc.) \\ 
    \midrule
    \multirow{3}*{\shortstack{BERT-Large \\ (335M)}}
    &CLS-FT$^1$ 
        & 77.7 & 94.6 & 93.7 & 74.9 & 75.8 & 24.7 & 70.5 & 68.3 & 69.0  & 72.5 \\
    &PET-FT
        & 77.2 & 91.1 & 93.5 & 70.5 & 73.6 & 17.7 & 67.0 & 80.8 & 75.0 & 73.1\\
    &\solution 
        & 77.8	& 96.4 & 97.4 & 72.7 & 75.5 & 17.1 & 65.6 & 81.7 & 76.0 &74.6\\
    \midrule[1pt]
    \multirow{3}*{\shortstack{GPT2-Med. \\ (345M)}}
    & CLS-FT
        & 71.0 & 73.2 & 51.2 & 65.2 & 72.2 & 19.2 & 65.8 & 62.5 & 66.0 & 63.1\\
    & PET-FT
        & 78.3 & 96.4 & 97.4 & 70.4 & 72.6 & 32.1 & 74.4 & 73.0 & 80.0  & 74.9 \\
    & \solution
    & 78.9 
    & 98.2 
    & 98.7 
    & 69.4 
    & 75.5 
    & 29.3 
    & 74.2 
    & 74.0 
    & 81.0 
    & 75.6 \\ 
    \bottomrule[1pt]
    \end{tabular}}
    \begin{tablenotes}
        \item[1] We report the same results taken from SuperGLUE~\cite{wang2019superglue}.
    \end{tablenotes}
    \end{threeparttable}
    }
    \caption{Fully-supervised performance on SuperGLUE development set. 
    }
    \label{tab:fully-supervised}
\end{table*}

%% file: 5_3_FewShot.tex
\subsection{Few-Shot Learning}\label{sec:few-shot}

While GPT-3 has shown decent few-shot learning potential with handcrafted prompts, it still struggles on some of the challenging tasks (e.g., natural language inference)~\cite{brown2020language}. We are motivated to study whether \solution can also improve the few-shot learning performance of pretrained models on challenging tasks.

\subsubsection{Setup}

\vpara{Few-shot Evaluation.}
The few-shot performance is sensitive to lots of factors (e.g., the order of training examples, random seed, and prompt patterns), and thus suffers from high variance~\cite{calibratebeforeuse,fantasticallyorderedprompts,revisitingfewsamplefinetune}. Therefore, the few-shot evaluation strategy should make sure that the improvements are indeed from an improved method instead of variance.
To this end, we follow the FewNLU evaluation procedure~\cite{zheng2021fewnlu} that has addressed and handled the issue.
Specifically, we use random data splits to perform model selection only on a small labeled set to prevent overfitting a large dev set.

\vpara{Dataset.}
We use the few-shot SuperGLUE (also known as FewGLUE) benchmark~\cite{schick2020small} and follow the setting in prior work \cite{zheng2021fewnlu} in terms of data split construction.
%

\vpara{Baseline and Hyper-parameter.} 
In few-shot learning, we again compare \solution with PET~\cite{schick2020small}, which was shown to outperform GPT-3 on some of the tasks.
Similar to~\cite{schick2020small}, we use ALBERT-xxLarge as the base model.
For hyper-parameters that are shared by PET and \solution (e.g., learning rate, maximum training step, evaluation frequency), we use the same search space for fair comparison.
Specifically, we search the learning rate in $\{1e-5, 2e-5\}$, the maximum training step in $\{250, 500\}$, and the evaluation frequency in $\{0.02, 0.04\}$.
    
\input{tables/few_main_results}
\input{tables/patternrobustness}

\vpara{Construction of Prompt Patterns.}
For PET, we use the same manual prompts reported by~\citet{schick2020small}.
When constructing prompt patterns for \solution, based on the same manual prompts as PET, we insert different numbers of continuous prompt tokens into different positions, thus formulating a number of pattern candidates.
We then select the best pattern for \solution using the validation strategy of FewNLU \cite{zheng2021fewnlu}.
We also conduct further analysis of the number and the position of continuous prompt tokens in \textsection\ref{sec:ablationnumber}.

\subsubsection{Main Results}
\vpara{Few-Shot Performance.}
Table~\ref{tab:fewshotmain} shows the main results of few-shot learning.
We find that, on ALBERT, \solution consistently outperform PET on average by more than 1 points.
It outperforms PromptTuning by more than 13 points.
It proves that by automatically learning continuous prompt tokens, the pretrained models can achieve better few-shot performance on NLU tasks.

\input{tables/few_prompt_number_location}

\subsubsection{Ablation Study}

\vpara{Type of Prompt Encoder}
Prior work~\cite{shin2020autoprompt} proposes to simply use an MLP as the prompt encoder, we perform further ablation analysis for prompt encoder selection, and results are shown in Table~\ref{tab:ablationpromptencoder}.
We consider LSTM, MLP, and EMB (i.e., we directly optimize the word embeddings without using additional parameters).
From the results, we can see that LSTM, MLP, and EMB all work as a prompt encoder.
Results show that both LSTM and MLP generally work well on these tasks, while EMB is unstable and can substantially under-perform the other two on some tasks (e.g,. WiC and CB).
To sum up, both LSTM and MLP could be taken into account when working on new tasks.

\input{tables/few_prompt_encoder}

\vpara{Location of Prompt Tokens}\label{sec:ablationnumber}
To study at which location to insert continuous prompt tokens, we perform experiments as Table~\ref{tab:ablation:number} shows.
From the results, we have the following findings.
\begin{enumerate}
    \item By comparing \#1 (or \#2) with \#3 (or \#4), we find that it would be better if we insert continuous prompt tokens at the location where it does not segment the sentences. For example, in case\#1, ``[P]'' breaks the completeness of sentence ``[Hypothesis]?'' while in case\#3, ``[P]'' is located between sentences.
    \item By comparing \#2 (or \#3) with \#4, we find that there's no special preference for placing on the edge or in the middle of the inputs.
    \item It is suggested to write a number of pattern candidates and then search over them for the best for each task.
\end{enumerate}

\vpara{Number of Prompt Tokens}\label{sec:ablationloacation}
We also study the influence of the number of prompt tokens and show the results in Table~\ref{tab:ablation:number}. 
By comparing \#3, \#6, \#7, and \#8, we can conclude that the number of prompt tokens has a great impact on the few-shot performance.
However, it is not that a larger number of prompt tokens would always be better.
We conjecture that it could be that due to the limited training data, it becomes difficult to learn the parameters when excessively increasing the number of continuous prompt tokens.
In practice, it is suggested to search for the best number of prompt tokens through model selection.

\subsubsection{Comparison with Discrete Prompt Search}

\input{tables/few_lmbff}

Prior work~\cite{gao2020making} proposed to automatically search discrete prompts and achieved better results than those of manual prompts. We now proceed to compare \ours with auto-searched discrete prompts.
For fair comparison, we follow the setting of LM-BFF \cite{gao2020making} to also conduct experiments on some of the GLUE tasks \cite{Wang2018GLUEAM} with RoBERTa-Large model \cite{liu2019roberta}.
Since the the evaluation protocols have large impacts on few-shot performance, we use the top-3 discrete prompts searched by LM-BFF and experiment with using only the discrete prompts and additionally applying \ours.
For \ours, the prompt patterns are constructed by concatenating the same discrete prompts as well as continuous prompts.
Results in Table~\ref{tab:few-shot:lmbff} show that additionally incorporating continuous prompts can further improve few-shot performance.
\ours is easy to be combined with existing discrete prompts, while further improving stability as discussed in Section~\ref{sec:stable}.

\subsection{Stabilizing Language Model Adaptation} \label{sec:stable}

In the above sections, we have shown that \ours improves over performance across multiple settings. Now we present results to demonstrate that \ours also stabilizes language model adaptation;
i.e., reducing the differences between different prompts.
As we have shown in Table~\ref{tab:motivation_lama}, manual prompts have a large impact on the performance.
When it comes to few-shot learning, the performance gap of different prompts is prominent due to the sensitivity of few-shot learning~\cite{zheng2021fewnlu}.
Results in Table~\ref{tab:fewshot:robustness} show that \solution improves the performance of the worst-performing patterns (e.g., P\#5), and achieves a smaller standard deviation over multiple patterns.
Compared to PET-FT, \solution increases the stability w.r.t. the choice of patterns.

On LAMA, we observe similar a phenomenon that while manual prompts often yield quite volatile results, appending trainable continuous prompts on top of the manual prompts can stabilize their performances, reducing the standard deviation from 10.1 to 0.46.

%% file: tables/few_main_results.tex
\begin{table*}
    \centering
    
    \resizebox{\linewidth}{!}{
    \begin{threeparttable}
    \begin{tabular}{l|cccccccccc}
    \toprule[1pt]
    Method
    & \makecell{BoolQ \\ (Acc.) }
    & \makecell{RTE \\ (Acc.) }
    & \makecell{WiC \\ (Acc.) }
    & \multicolumn{2}{c}{\makecell{CB \\(Acc.) \qquad (F1.)}}
    & \multicolumn{2}{c}{\makecell{MultiRC \\ (F1a.) \qquad (EM.)}}
    & \makecell{WSC \\ (Acc.) }
    & \makecell{COPA \\ (Acc.) }
    & Avg  \\
    \midrule
    Prompt Tuning
    & 58.47	\std1.00
    & 54.42	\std3.05
    & 52.74	\std2.36
    & 75.45	\std2.25
    & 67.73	\std5.70
    & 59.28	\std4.73
    & 15.03	\std4.11
    & 74.04	\std2.99
    & 61.50	\std4.36
    & 58.56
    \\
    
    PET-FT
    & 76.70 \std1.85
    & 72.83 \std1.30
    & 53.87 \std4.47
    & 84.38 \std4.47
    & 62.56 \std7.66
    & 76.51 \std1.52
    & 36.46 \std2.13
    & 80.05 \std2.53
    & 81.75 \std4.03
    & 70.74
    \\
    \solution
    & 76.55 \std2.68
    & 63.27 \std3.63
    & 55.49 \std1.21
    & 88.39 \std3.72
    & 84.24 \std5.15
    & 75.91 \std1.74
    & 38.01 \std0.78
    & 78.85 \std1.76 
    & 85.25 \std3.30
    & 71.81
    \\
    \bottomrule[1pt]
    \end{tabular}
\end{threeparttable}
}
\caption{{The few-shot performance of PET~\cite{schick2020small}, Prompt Tuning~\cite{Lester2021ThePO} and our \solution over seven tasks based on ALBERT. 
Each result is averaged over 4 runs with different data splits. 
Results show that \solution consistently improves average few-shot performance by more than 1 point compared to PET and by more than 13 points compared to Prompt Tuning.}}
\label{tab:fewshotmain}
\end{table*}

%% file: tables/patternrobustness.tex
\begin{table*}[t]
    \centering
    \small
    \resizebox{0.7\linewidth}{!}
    {
    \begin{tabular}{c|l|cccccc|c}
    \toprule[1pt]
         
         & Method
         & P\#0 
         & P\#1 
         & P\#2 
         & P\#3 
         & P\#4 
         & P\#5  
         & STD \\
         \midrule
        \multirow{4}*{\tabincell{c}{FSL\\(BoolQ)}}
         & \multirow{2}*{PET-FT} 
         & 77.10 
         & 67.96
         & 74.14
         & 72.48
         & 71.77
         & 60.86
         & \multirow{2}*{5.68} \\
         &
         & \std 2.21
         & \std 2.69
         & \std 1.38
         & \std 4.31
         & \std 2.56
         & \std 3.99
         & 
         \\
         &
        \multirow{2}*{\solution}  
         & 75.41
         & 75.11
         & 73.43
         & 71.35
         & 71.31
         & 65.86
         & \multirow{2}*{3.52}
         \\
         &
         & \std 3.09 
         & \std 1.61
         & \std 2.60
         & \std 4.57
         & \std 8.58
         & \std 3.80
         & \\
    %
         \midrule
        \multirow{2}*{\tabincell{c}{LAMA\\(P17)}}
        & \multirow{1}*{MP}  
        & 31.3	
        & 19.8	
        & 31.4	
        & 51.1	
        & 34.0	
        & 32.7	
        & 10.1 \\
        &
        \multirow{1}*{\solution} 
        & 57.8
        & 57.8
        & 58.1
        & 58.1
        & 58.9
        & 58.7
        & 0.46 \\
    \bottomrule[1pt]
    \end{tabular}
    }
    
    \caption{{Upper table: Few-shot learning (FSL) of PET and \solution in terms of each pattern on SuperGLUE with ALBERT; Lower table: Manual prompt (MP) and \solution performance on LAMA-P17 with BERT-base-cased. For each column, \solution and compared methods share the same manual prompts, while \solution additionally concatenates continuous prompt tokens. We report the standard deviation over multiple results of different patterns. Results show that \solution achieves smaller standard deviation, proving that \solution can improve stability w.r.t. the choice of discrete patterns.}}
    \label{tab:fewshot:robustness}
\end{table*}

%% file: tables/few_prompt_number_location.tex
\begin{table*}
    \centering
    \resizebox{\linewidth}{!}{
    \begin{tabular}{l|l|ccc|cc|c}
    \toprule[1pt]
         ID 
         & Prompt Patterns of \solution
         & Seg.
         & Pos.
         & \#[P]
         & Acc. & F1. & Avg.\\
    \midrule
    1
    & [Premise] Question: [Hypothesis] [P] ? Answer: [M].
    & Yes
    & Mid
    & 1
    & 87.95
    & 76.70 & 82.33\\
    2
    & [Premise] Question [P]: [Hypothesis] ? Answer: [M].
    & Yes
    & Mid
    & 1 
    & 88.39
    & 78.57 &   83.48
\\
    3
    & [Premise] Question: [Hypothesis] ? [P] Answer: [M].
    & No
    & Mid
    & 1
    & 89.29
    & 79.86 & 84.58

\\
    4
    & [Premise] [P] Question: [Hypothesis] ? Answer: [M].
    & No
    & Miid
    & 1
    & 89.73
    & 82.15 & 85.94
    \\
    5 
    & [Premise] Question: [Hypothesis] ? Answer: [M]. [P] 
    & No
    & Edge
    & 1
    & 87.50
    & 83.39 & 85.45
\\
    6
    & [Premise] Question: [Hypothesis] ? [P][P] Answer: [M].
    & No
    & Mid
    & 2
    & 88.39
    & 84.74 & 86.57
\\
    7 
    & [Premise] Question: [Hypothesis] ? [P][P][P][P] Answer: [M].
    & No
    & Mid
    & 4
    & 88.39
    & 85.14 & 86.76
\\
    8
    & [Premise] Question: [Hypothesis] ? [P][P][P][P][P][P][P][P] Answer: [M].
    & No
    & Mid
    & 8
    & 83.48
    & 73.32 & 78.40
\\
    \bottomrule[1pt]
    \end{tabular}}
    \caption{{The few-shot performance of \solution on the CB task on ALBERT with different prompt patterns. ``Seg.'' means whether the inserted prompt tokens segment complete sentences. ``Pos.'' indicates inserting the prompt tokens at the edge or in the middle of the inputs.
    ``[P]'' is continuous prompt token. ``[M]'' is the mask token.}}
    \label{tab:ablation:number}
\end{table*}

%% file: tables/few_prompt_encoder.tex
\begin{table}[ht]
    \centering
    \small
    \resizebox{\linewidth}{!}{
    \begin{tabular}{l|ccc}
    \toprule[1pt]
         Task  &  LSTM & MLP & EMB \\
    \midrule
        
         \multirow{1}*{WiC-ACC}    
         & 56.27 \std1.54
         & 55.25 \std3.09
         & 53.96 \std3.23
         \\
         \midrule
         \multirow{1}*{CB-ACC.}
         & 81.70 \std 7.49 & 88.39 \std3.72 & 82.59\std3.69 \\
         \multirow{1}*{CB-F1.}
         & 77.41 \std9.15 & 84.24\std5.15 & 67.27\std6.78 \\
         \midrule
         \multirow{1}*{BoolQ-ACC.}
         & 75.41\std 3.09 & 76.46\std2.84 & 76.87\std1.69 \\
    \bottomrule[1pt]
    \end{tabular}}
    \caption{{The few-shot performance on WiC, CB and BoolQ tasks with ALBERT using different prompt encoders. Results show that both LSTM and MLP generally work well on these tasks, while EMB is unstable and can substantially under-perform the other two on some tasks (e.g,. WiC and CB). ``EMB'' means using an identity mapping for the prompt encoder.}}
    \label{tab:ablationpromptencoder}
\end{table}

%% file: tables/few_lmbff.tex
\begin{table}[]
    \centering
    \resizebox{0.7\linewidth}{!}{
    \begin{tabular}{l|cc}
    \toprule[1pt]
      Task  &  LM-BFF (Auto) & \ours\\
     \midrule
     SST-2 & 92.89 & 92.78\\
     MNLI & 57.53 & 58.70 \\
     MRPC & 68.26 & 69.49 \\
     \bottomrule[1pt]
    \end{tabular}}
    \caption{{
    Few-shot performance of automatically searched prompts and \ours. 
    We evaluated LM-BFF (Auto) using the reported top-3 searched patterns under our evaluation procedure. 
    \ours also uses the same discrete prompts, in concatenation with continuous prompts.
    Results show that \ours can be effectively combined with existing discrete patterns and achieve further performance improvement.
    }}
    \label{tab:few-shot:lmbff}
\end{table}

%% file: 2_related.tex
\section{Related work}

\vpara{Language Model Prompting.}
GPT-3 \cite{brown2020language} uses in-context examples \cite{liu2021makes,zhao2021calibrate} as a way of prompting to transfer knowledge from pretraining to downstream tasks. \citet{schick2020small} proposed to use cloze patterns, which removes the constraint that the masked token is the last token of the sentence. This further minimizes the gap between pretraining and downstream tasks. To improve prompting for NLU, recent works have proposed methods to automatically search for high-performing prompts by mining the training corpus \cite{jiang2020can}, gradient-based search \cite{shin2020autoprompt}, or using pretrained generative models \cite{gao2020making}. Our approach is different from these prior works in that we resort to using continuous prompt embeddings, which are found to be complementary to discrete prompts in our experiments.


Recently, some concurrent works also proposed the use of continuous prompts. Prefix-tuning ~\cite{li2021prefix} adds continuous prompts at the beginning of the sequence for each layer. In contrast to our work, prefix-tuning targets natural language generation tasks.

In the area of NLU, a few concurrent methods were proposed based on continuous prompts, focusing on improving knowledge probing \cite{Qin2021LearningHT,Zhong2021FactualPI}. \citet{Lester2021ThePO} showed that with large pretrained models, only tuning continuous prompts with a frozen language model achieves comparable performance to full-model tuning.

Compared to these concurrent works on NLU,
P-Tuning reaches a unique conclusion that continuous prompts improve performance and stabilize training with either frozen or tuned models under both the few-shot and fully-supervised settings. For example, no concurrent works have shown that continuous prompts can improve performance with a tuned language model. Technically, P-Tuning also has a few unique designs such as using hybrid continuous-discrete prompts and employing a prompt encoder.


\vpara{Knowledge in Language Models.}
Self-supervised~\cite{liu2020self} pre-trained language models~\cite{han2021pre} including GPT~\cite{radford2019language}, BERT~\cite{devlin2018bert}, XLNet~\cite{yang2019xlnet}, RoBERTa~\cite{liu2019roberta} have been observed to learn not only contextualized text representations but also linguistic and world knowledge. 
~\cite{hewitt2019structural} demonstrates that contextualized representations produced by language models can form a parse tree in the embedding space. ~\cite{vig2019multiscale, clark2019does} look into the multi-head attention patterns within transformers and discover that certain attention heads may correspond to some grammatical functions, including co-reference and noun modifiers.
LAMA~\cite{petroni2019language,petroni2020context} propose the LAMA task that leverages cloze tests to predict the fact triples of knowledge bases to examine language model's ability of memorizing facts with answers in the single-token format. In~\cite{wang2020language}, the authors investigate the attention matrices to find evidence about knowledge triples contained in the context. ~\cite{jiang2020x} develops a multi-token fact retrieval dataset based on LAMA.

%% file: 6_conclusion.tex
\section{Conclusions}

In this paper, we present a method \ours that uses continuous prompts in concatenation with discrete prompts. \ours improves performance and stabilizes training for pretrained language model adaptation. \ours is effective with both tuned and frozen language models under both the few-shot and fully-supervised setings.
